\documentclass[letterpaper]{article}
\usepackage{iccc}

\usepackage{times}
\usepackage{helvet}
\usepackage{courier}
\usepackage[utf8]{inputenc}

\usepackage{graphicx}

\usepackage{url}
\usepackage{tikz}
\usepackage{tikz-dependency}
\usepackage{subcaption}
\usepackage{tabularx}
\usepackage{booktabs}
\usepackage{float}
\usepackage{multirow}

\title{When a Computer Cracks a Joke: Automated Generation of Humorous Headlines}
\author{Khalid Alnajjar\\
Faculty of Arts\\
University of Helsinki\\
khalid.alnajjar@helsinki.fi
\And
Mika Hämäläinen\\
Faculty of Arts\\
University of Helsinki\\
mika.hamalainen@helsinki.fi
}
\begin{document} 
\maketitle
\begin{abstract}
\begin{quote}
Automated news generation has become a major interest for new agencies in the past. Oftentimes headlines for such automatically generated news articles are unimaginative as they have been generated with ready-made templates. We present a computationally creative approach for headline generation that can generate humorous versions of existing headlines. We evaluate our system with human judges and compare the results to human authored humorous titles. The headlines produced by the system are considered funny 36\% of the time by human evaluators.
\end{quote}
\end{abstract}

\section{Introduction}

Humor, while showcased by a wide spectrum of species in the animal kingdom, has a subcategory that is exclusive to the human being. Verbal humor can only exist in the presence of language, and its generation by computational means is far from trivial.

Humor is effectively a perceiver dependent phenomenon. Nothing can be inherently funny, but humor is perceived and appraised by the mind of a human perceiving it. And ultimately, the perceived humor, if accepted as such by the listener, elicits an emotional response accompanied with a vocal response of laughter that originates from our ancestors, the species before homo sapiens (cf. \citeauthor{ross2010evolution} 2010).

Our paper focuses on generating humor in news headlines. This NLG task is not of the traditional sort, where conveying factual information is the uttermost goal of the system, but rather the affective content of the message is taken into the primary focus of the study.

Automated news generation is a flourishing field with new research being published in a timely manner. This is reflected by the number of recent publications on the topic \cite{nesterenko-2016-building,yao-etal-2017-content}. Quite often, however, as the generated news has to cater for the purpose of communicating facts the question of creativity is set aside. In such a context, there is a trade-off between creativity and communicativity (see \citeauthor{nlp4convai} 2019).

Whereas creative headline generation is not a new domain to computational creativity, with quite some existing publications on the topic \cite{lynchevery,gatti2015slogans,alnajjar2019no}, we aim to intertwine headlines, creativity and humor by proposing a novel method for humor generation that is reasoned by the existing theories on humor.

Our approach alters a word in an existing headline for a humorous effect. We evaluate the method proposed in this paper quantitatively with human judges. We take the different constituents of humor in consideration in the evaluation to uncover the relation of each feature to the humor produced by our system.

\section{Related Work}

Humor has received some interest in the past for more than a decade \cite{ritchie2005computational,hong2009automatically,valitutti2013let,costa2015reality}. We dedicate the remaining of this section to describing some of the most recent work conducted on the topic.

Pun generation with a neural model language model is one of the most recent efforts on humor generation \cite{yu2018neural}. Their approach consists of training a conditional language model an using a beam search to find sentences that can support two polysemous meanings for a given word. In addition they train a model to highlight the different meanings of the word in the sentence. Unfortunately, they evaluate their system on human evaluators based on three quantitative metrics: fluency, accuracy and readability, none of which tells anything about how funny or apt the puns were.

\citeauthor{inlg} \shortcite{inlg} present a genetic algorithm approach for generating humorous and satirical movie titles out of existing ones. Their method works on a word level replacement and aims for low semantic similarity of the replacement word with the original word to maximize surprise and high similarity with Saudi Arabia to maximize coherence. They consider pun as one of the fitness functions of the genetic algorithm, but the output is not strictly limited to puns. On top the genetic algorithm, they train an RNN model that learns from the genetic algorithm and real people.

Surprise is also one of the key aspects of a recent pun generator \cite{he-etal-2019-pun}. They model surprise as conditional probabilities. They introduce a local surprise model to assess the surprise in the immediate context of the pun word and a global surprise to assess the surprise in the context of the whole text. Their approach retrieves text from a corpus based on an original word - pun word pair. They do a word replacement for local surprise and insert a topic word for global surprise.

An approach building on humor theories is that of \citeauthor{neverwinter} \shortcite{neverwinter}. The theories are used in feature engineering. They learn templates and metrical schemata from jokes rated by people with a star rating. They embrace more traditional machine learning techniques over neural networks, which has the advantage of a greater interpretability of the models.

Humor has also been tried to recognize automatically in the past. One of such attempts is focuses on extracting humor anchors, i.e. words that can make text humorous, automatically \cite{yang2015humor}. A similar humor anchor based approach is also embraced by \citeauthor{cattle-ma-2018-recognizing} \shortcite{cattle-ma-2018-recognizing}. Both of the approaches rely on feature engineering basing on humor theories. Recently LSTM models have been used for the task of humor detection with a different rates of success \cite{cai-etal-2018-sense,sane-etal-2019-deep,zou-lu-2019-joint}.

\subsection{Humor}

Humor is an inherent part of being a human and as such it has provoked the interest of many researchers in the past to formulate a definition for it (see \cite{krikmann2006contemporary}). \citeauthor{koestler1964act} (1964) sees humor as a part of creativity together with discovery and art. In his view, what is characteristic to humor in comparison to the other two constituents of creativity, is that its emotional mood is aggressive in its nature. He calls bisociation in humor the collision of two frames of reference in a comic way.

\citeauthor{raskin1984semantic} \shortcite{raskin1984semantic} presents a theory that is not too far away from the previously described one in the sense that in order for text to be humorous, it has to be compatible with two different scripts. The different scripts have to be somehow in opposition, for example in the sense that one script is a real situation and the other is not real. 

In \citeauthor{attardo1991script} \shortcite{attardo1991script} humor is seen to consist of six hierarchical knowledge resources: language, narrative strategy, target, situation, logical mechanism and script opposition. As in the previous theories, the incongruity of two possible interpretations is seen as an important aspect for humor. An interesting notion that we will take into a closer examination is that of target. According to the authors it is not uncommon for a joke to have a target, such as an important political person or an ethnic group, to be made fun of.

Two requirements have been suggested in the past as components of humor in jokes: surprise and coherence (see \cite{brownell1983surprise}). A joke will then consist of a surprising element that will need to be coherent in the context of the joke. This is similar to having two incongruous scripts being simultaneously possible.

\citeauthor{veale2004incongruity} \shortcite{veale2004incongruity} points out that the theories of \citeauthor{raskin1984semantic} \shortcite{raskin1984semantic} and \citeauthor{attardo1991script} \shortcite{attardo1991script} entail people to be forced into resolution of humor. He argues that humor should not be seen as resolution of incompatible scripts, but rather as a collaboration, where the listener willingly accepts the humorous interpretation of the joke. Moreover, he argues that while incongruity contributes to humor, it does not alone constitute it.

\section{Generating Humorous Headlines}

In their work, \citeauthor{hossain-etal-2019-president} \shortcite{hossain-etal-2019-president} identified several ways people altered news headlines to be humorous. In our method, we aim to model the following ones of their findings: the replacement forms a meaningful n-gram, the replacements are semantically distant, the replacement makes a strong connection with the entity of the headline and belittles an entity or a noun and the replacement creates incongruity. We see the n-gram finding in a broader way of the replacement being compatible with the the existing script (context). The semantic distance is seen as an index of surprise, and the connection between the entity is assimilated with the target of the joke.

The findings we are not focusing on in this paper are that the replacements are sarcastic, suppress tension or have a setup and punchline. The first two are left out as assessing them computationally is a task worth of a paper on their own right, and the third one is left out as it focuses on a particular kind of humor. However, the punchline structure might emerge from the other features being modelled although not explicitly taken into consideration.

In addition to the findings described above, we take the concreteness of the replacement word into account. The reason for this that concrete words are more likely to provoke mental images (see \cite{concretepoems}). In fact, we could see this in the humorous training dataset by \citeauthor{hossain-etal-2019-president} \shortcite{hossain-etal-2019-president}, where 90\% of the most humorous replacement words were concrete as opposed to only 75\% of the least humorous replacement words being concrete.

For the above experiment and the rest of the paper, we use the lexicon of 40k common English words that has a concreteness score from 1 to 5 assigned \cite{brysbaert2014concreteness}. If the score assigned with the word is greater or equal to 3, we consider it concrete. The concreteness is evaluated by lemmatizing the word with spaCy~\cite{spacy2} if it does not exist in the lexicon.

\subsection{Modelling Humor}

Our system operates by taking an existing headline from the corpus of altered headlines \cite{hossain-etal-2019-president}. This corpus has been syntactically parsed by us by using spaCy \cite{spacy2}, and it has been tagged for the words that should be replaced by its original authors. For a selected headline, our system tries to find suitable humoristic replacement words.

\begin{figure*}[!htb]
\center{\includegraphics[width=\textwidth]
{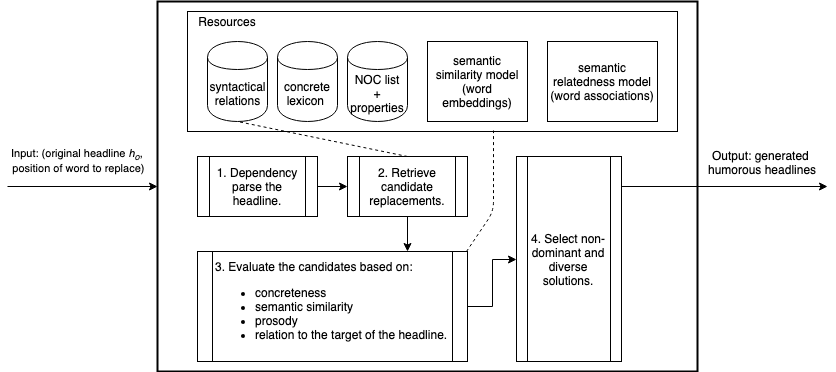}}
\caption{\label{fig:diagram2} A diagram visualizing the process of humor generation.}
\end{figure*}

We assess the different potential humorous replacements in terms of multiple parameters, which are prosody, concreteness, semantic similarity of the replacement to the original word and the semantic relatedness of the replacement to negative words describing the target. In this section, we explain how the individual parameters are modelled. An overall view of our method is depicted in Figure \ref{fig:diagram2}.

For prosody, we look at the sound similarity between the original word and the replacement. We assess this in terms of full rhymes, assonance, consonance and alliteration. These are implemented with rules. As the written form of English is notoriously deviant from the phonation, we use eSpeak-ng\footnote{UK English voice, https://github.com/espeak-ng/espeak-ng} to produce IPA transcription for the words the prosody of which is being assessed.

For concreteness we use the values provided in \citeauthor{brysbaert2014concreteness} (2014) to score the concreteness of the replacement word. And for semantic similarity we use the pretrained word embeddings from \citeauthor{bojanowski2017enriching} (2017). We use the semantic similarity to assess surprise, in other words, we want to minimize the similarity of the replacement word to the original.

To measure how a new replacement connects to the word selected to be the target of the joke in the headline, a target must first be found. We consider recognized entities in the headline as the potential targets. In case no entities were recognized, we use the subjects in the headline. If neither of them existed, nouns in the headline are treated as target. Out of the list of targets, a random target $t$ is picked to focus on. For this target $t$, we retrieve words that are related to it to act as descriptive words revealing potential attributes to make fun of. We employ two resources to obtain such knowledge regarding the selected target:
\begin{enumerate}
    \item The Non-Official Characterization (NOC) list~\cite{veale-2016-round} which contains information about more than 1000 well-known characters (e.g. \textit{Donald Trump} and \textit{Kim Jung-un}) and their expanded stereotypical properties supplied by~\cite{alnajjar2017expanding} (e.g. \textit{Donald Trump}: [wealthy, successful, greedy, aggressive, \ldots etc]).
    \item A semantic relatedness model built from word associations collected from a web text corpus ukWac\footnote{\url{https://wacky.sslmit.unibo.it/doku.php?id=corpora}}, following the approach described in Meta4meaning~\cite{meta4meaning981}. We chose to base our relatedness model on a web-based corpus instead of a news-based one to favor discovering related words from various domains, which would be perceived as more humorous.
\end{enumerate}

If the target $t$ is an entity, we search the first resources (i.e. the NOC list and the expanded properties) to collect its top $k$ stereotypical properties. In case no available knowledge regarding the entity existed, we attempt to acquire the top $k$ related words to the rest of the potential targets (subjects and nouns, respectively) using the second resource (i.e. the semantic relatedness model). In our case, we empirically set $k$ to 100 to allow diversity and reduce noisy relations, while ensuring the descriptiveness of the words to the target.

To be able to place the target in a humorous light, we only regard the descriptive words that describe it negatively, which is determined by employing a polarity classifier provided by~\citeauthor{akbik2018coling} (2018). Lastly, the connection of the replacement word to the target is assessed based on the semantic relatedness between the replacement word and the target's negative descriptions. We desire to maximize such connections to encourage replacements that are associated with the target from a negative angle.

\subsection{Generation and picking out the best candidate}

We use the Humicroedit dataset of headlines published by \citeauthor{hossain-etal-2019-president} (2019) as the source of original headlines. Furthermore, the dataset contains edits performed by humans to make the headlines humorous along with a score indicating how humorous they were when perceived by other people on a scale from 0 to 3. The motivation for using this dataset is that the editors were required to make a single change to either a verb or a noun in the headline to make it humorous, which focuses the scope when modeling such a process computationally.

In our generation method, we only consider headlines where the original word that is selected to be replaced is parsed as either a noun or a verb using spaCy and is a single token (i.e. ignoring cases such as ``Illegal Immigrants''). The rational behind is to reduce misparsing errors and concentrate on a single-word changes. 

\begin{figure}
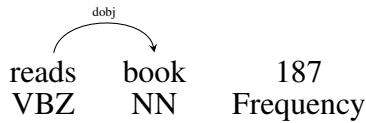

\centering
\Large
\begin{dependency}[theme = simple]
 \begin{deptext}[column sep=1em]
  reads \& book \& 187 \\
  VBZ \& NN \& Frequency\\
 \end{deptext}
 \depedge{1}{2}{dobj}
\end{dependency}
\caption{An example of a syntactical relation in the repository of grammatical relations~\cite{khalid_alnajjar_2018} along with its frequency.}
\label{fig:relation-example}
\end{figure}

For an original headline $h_{o}$ with its selected word to be replace $w_{o}$, our method converts it into a humorous one $h_{h}$ by replacing $w_{o}$ with another word  $w_{h}$ as follows. It begins by acquiring replacement candidates $C$ that fit the syntactical position of the selected word $w_{o}$ by querying a massive syntactical repository of grammatical relations that have a frequency greater than 50
in a web-based corpus~\cite{khalid_alnajjar_2018} (see Figure~\ref{fig:relation-example} for an example of a grammatical relation in the repository). By considering candidates that are apt to the existing syntactical relations in the headline, we ensure that the new replacement has syntactic cohesion and suits the grammatical context.

To illustrate how the method works, let's consider the headline $h_{o}=$ ``City halls and landmarks turn green in support of Paris \textit{climate} deal'' as an example, where the word to replace $w_{o}=$ \textit{climate}. After parsing this headline, we find that the to-be-replaced word $w_{o}$ is a noun (NN) and has a dependency (compound) on the word \textit{deal} (NN). We query the syntactical repertory to find potential replacements that suit this relation, which yields 58 candidate replacements ($C = $\{`loan', `business', `cash', `oil', `holiday', `peace', `content', `drug' \ldots etc\}).

In the next phase, the method removes the original word $w_{o}$ from the candidates if it existed and prunes out any candidate word in $C$ that is not identified as concrete (i.e. having a concreteness score greater or equal to 3 based on~\cite{brysbaert2014concreteness}). As a result, candidate words such as `peace' and `content' in the earlier example are removed resulting in a total of 34 candidates. If there is more than 500 replacement candidates (e.g. in situations where the token to replace is a verb and is the root of the phrase), we randomly select 500 candidates in $C$ to be examined. This is performed to reduce the search space that the method will traverse and to efficiently discover local optimal solutions as there is no particular global optimal solution for the task we are addressing.

Replacement candidates are then evaluated on the four humour aspects we are modeling, which are 1) prosody, 2) concreteness score, 3) inverted (i.e. minimized) semantic similarity between the original word $w_{o}$ and the candidate $c$, and 4) the semantic relation between the candidate $c$ and the negative words of the selected target $t$. As we are dealing with multiple criteria for modeling humour, we adopt a non-dominant multi-objective sorting approach~\cite{10.1007/3-540-45356-3_83} to find and select candidates in the Pareto front. Additionally, applying a non-dominant sorting for creative tasks (e.g. generating humour) increases the chances of finding balanced and diverse solutions that are more likely to be deemed good~\cite{alnajjar2018slogans}.

Applying the evaluation and the non-dominant sorting on the example headline, the method highlights candidates such as `cash', `meal', `drug' to be chosen as replacements. For the same example, the original word \textit{climate} was replaced with \textit{marijuana} by a human editor in the Humicroedit dataset. Interestingly, \textit{marijuana} is a \textit{drug} and our method was able to suggest it.

\section{Results and Evaluation}

To evaluate our method, we randomly select 83 headlines from the Humicroedit dataset that meet our criteria specified earlier. For each headline, we request our method to produce humorous alternatives, ranked by the non-dominant sorting, out of which we randomly select 3 to be evaluated from the top humorous headlines.

\begin{table*}[!ht]
\centering
\begin{tabular}{|l|c|c|}
\hline
\textbf{Humorous headline by our system} & \multicolumn{1}{l|}{\textbf{Original word}} & \multicolumn{1}{l|}{\textbf{Human replacement}} \\ \hline
\begin{tabular}[c]{@{}l@{}}Thieves carry out elaborate van heist to steal millions\\  in \textbf{cereal}, Swiss police say\end{tabular} & cash & blouses \\ \hline
Trump \textbf{eats} the wrong Lee Greenwood on Twitter & tags & woos \\ \hline
\begin{tabular}[c]{@{}l@{}}'I was very angry' at Trump, says Myeshia Johnson,\\  widow of fallen \textbf{sock}\end{tabular} & soldier & cake \\ \hline
\begin{tabular}[c]{@{}l@{}}Trump Tried To \textbf{Climb} Heather Heyer’s Mother \\ During Funeral: ‘I Have Not And Now I Will Not’\\  Talk To Him\end{tabular} & call & proposition \\ \hline
\begin{tabular}[c]{@{}l@{}}U.S. says Turkey is helping ISIS by \textbf{Combing} \\ Kurds in Syria\end{tabular} & bombing & feeding \\ \hline
\end{tabular}
\caption{Examples of generated headlines.}
\label{tab:headline_examples}
\end{table*}

Table \ref{tab:headline_examples} shows some of the headlines generated by our approach. The humorous replacement word is marked in bold. The original word and the replacement word suggested by a human from the corpus are shown in their respective columns.

We conduct our evaluation on Figure-Eight\footnote{https://www.figure-eight.com/}, which is a crowd-sourcing platform that assigns paid reviewers for tasks such as questionnaires. We evaluate all the 3 variations produced by our system for the 83 headlines, showing the original headline as well. In addition, we evaluate the human edits for the same headlines from the dataset. The reviewers were not told they were evaluating computer generated humor, as the mere knowledge of a computer being an author of a creative artefact is known to provoke a bias towards seeing the generated output in a more negative light (see \cite{colton2012computational}).

We asked five people to rate the headlines based on the following questions:

\begin{enumerate}
    \item The altered headline is humorous.
    \item The altered word is surprising.
    \item The altered word fits into the headline.
    \item The altered word is concrete.
    \item The joke of the headline makes fun of a person or a group of people (also known as the target of the joke).
    \item The altered word shows the target in a negative light.
    \item The altered word is a pun of the original word.
\end{enumerate}

We evaluate the first two questions on the scale form 0 to 3 (\textit{Not funny}, \textit{Slightly funny}, \textit{Moderately funny} and \textit{Funny}. Or surprising in the case of the Q2) similarly to the questions for humor in ~\citeauthor{hossain-etal-2017-filling} (2017). The rest of the questions are presented as yes/no questions. The sixth question is only visible if the fifth question has been answered to affirmatively.

\begin{figure}[!htb]
\center{\includegraphics[width=7.5cm]
{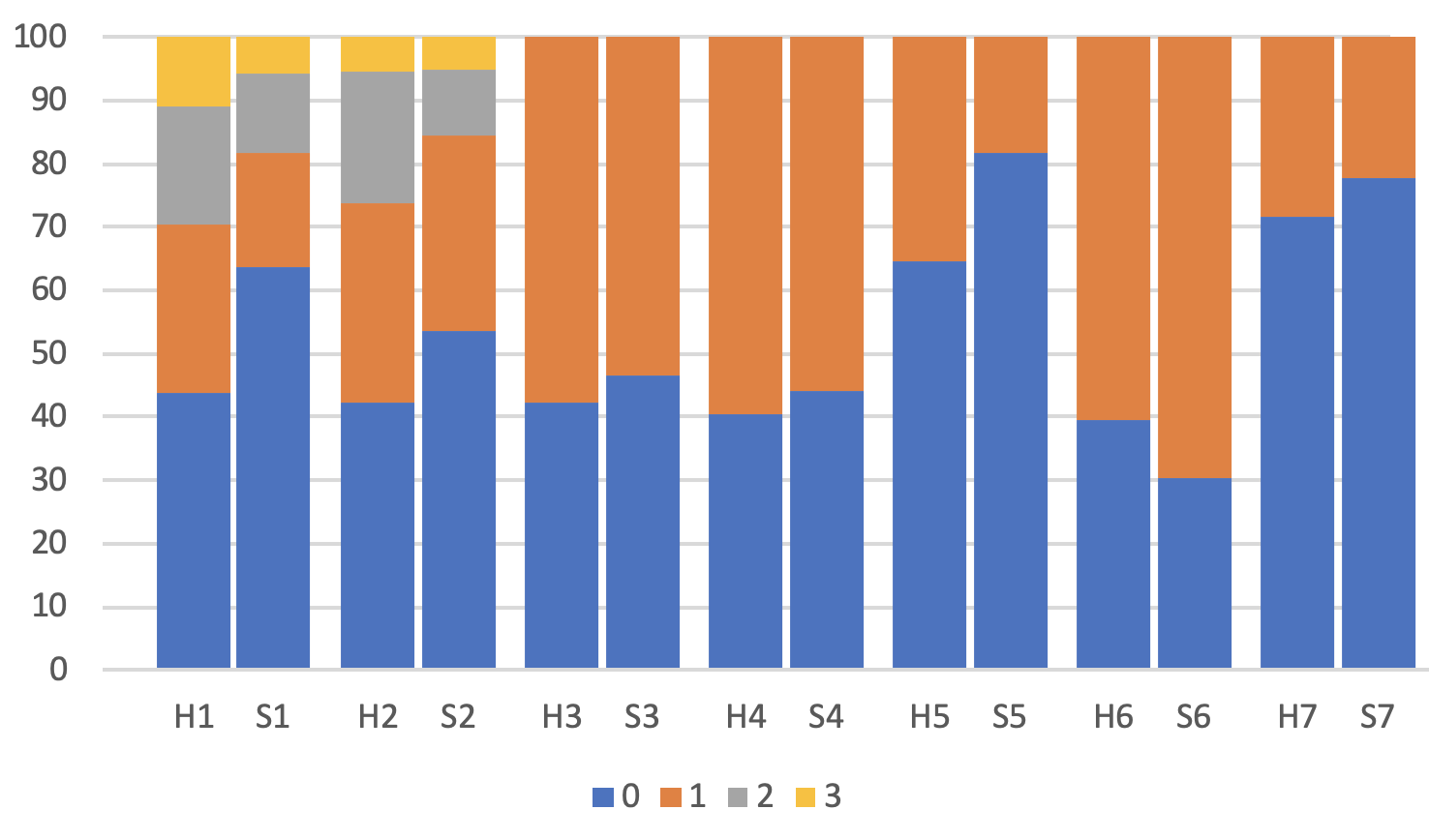}}
\caption{\label{fig:dist-plot}Percentages for each evaluation question. H marking human authored headlines, and S computer authored ones.}
\end{figure}

Figure \ref{fig:dist-plot} shows the percentages of the result for each question from the human evaluation. The results for the human edited titles (H) and the ones produced by our method (S) are shown side by side. From the question 3 onward, 0 marks negative and 1 affirmative answer.  All in all, our system scored slightly lower on the questions than real people, which is to be expected due to the difficulty of the problem. However, our system got slightly better results in the question number 6, which means, that when the system had recognizably picked a target, it managed to convey negativity towards the target on a level comparable to a real human.

In terms of humor, our system managed to produce at least slightly humorous headlines 36\% of the time, whereas people produced at least slightly humorous headlines 56\% of the time. In comparison, for a recent pun generator, \cite{he-etal-2019-pun} report a success rate of 31\% for their system according to a human evaluation, to put our results in a computational perspective.

\begin{table*}[ht!]
\resizebox{\textwidth}{!}{%
\begin{tabular}{|c|c|c|c|c|c|c|c|c|c|c|c|c|c|c|}
\hline
\multirow{2}{*}{} & \multicolumn{2}{c|}{\textbf{Q1}} & \multicolumn{2}{c|}{\textbf{Q2}} & \multicolumn{2}{c|}{\textbf{Q3}} & \multicolumn{2}{c|}{\textbf{Q4}} & \multicolumn{2}{c|}{\textbf{Q5}} & \multicolumn{2}{c|}{\textbf{Q6}} & \multicolumn{2}{c|}{\textbf{Q7}} \\ \cline{2-15} 
 & \multicolumn{1}{l|}{$\mu_{x}$} & \multicolumn{1}{l|}{$SD$} & \multicolumn{1}{l|}{$\mu_{x}$} & \multicolumn{1}{l|}{$SD$} & \multicolumn{1}{l|}{$\mu_{x}$} & \multicolumn{1}{l|}{$SD$} & \multicolumn{1}{l|}{$\mu_{x}$} & \multicolumn{1}{l|}{$SD$} & \multicolumn{1}{l|}{$\mu_{x}$} & \multicolumn{1}{l|}{$SD$} & \multicolumn{1}{l|}{$\mu_{x}$} & \multicolumn{1}{l|}{$SD$} & \multicolumn{1}{l|}{$\mu_{x}$} & \multicolumn{1}{l|}{$SD$} \\ \hline
score & 0,99 & 0,62 & - & - & - & - & - & - & - & - & - & - & - & - \\ \hline
human & 0,97 & 0,49 & 0,89 & 0,41 & 0,58 & 0,23 & 0,6 & 0,2 & 0,36 & 0,23 & 0,61 & 0,41 & 0,28 & 0,18 \\ \hline
max & 0,97 & 0,45 & 0,98 & 0,36 & 0,69 & 0,19 & 0,7 & 0,15 & 0,33 & 0,17 & 0,86 & 0,34 & 0,35 & 0,17 \\ \hline
avg & 0,6 & 0,28 & 0,67 & 0,23 & 0,53 & 0,15 & 0,56 & 0,11 & 0,19 & 0,12 & 0,6 & 0,31 & 0,22 & 0,11 \\ \hline
min	& 0,28	& 0,27	&	0,37	&	0,24	&	0,36	&	0,18	&	0,4	&	0,16	&	0,06	&	0,12	&	0,27	&	0,39	&	0,1	&	0,11\\ \hline
\end{tabular}%
}
\caption{Mean and standard deviation of altered headlines by humans and our method.}
\label{tab:mean-std}
\end{table*}

Table \ref{tab:mean-std} shows the results form another perspective. The \textit{score} row shows the results for human authored titles in the original publication \cite{hossain-etal-2017-filling}, whereas the \textit{human} row shows the results for the very same titles in our evaluation. The \textit{max} shows the average of the best scoring generated headline out of the 3 ones produced for each original headline, and \textit{min} shows the average of the worst headline in the triplets. \textit{Avg} is the average of the scores for all the generated headlines.

By looking at the results this way, we can see that at best, our method can produce humor comparable to real humans in the scale of funniness, with a higher amount of surprise, better aptness of the replacement word to the context, higher level of concreteness, higher negativity towards the target and higher level of puniness, falling shorter only in the case of having a perceivable target for the joke in the headline. Focusing on the best scoring individuals might sound like giving too good a picture of the performance of the system, however, they set the upper boundary for the performance of the system. This being said, with the exact same method, better results could be obtained in the future by developing a better way for ranking the humorous headline candidates output by the system.

By considering the headlines produced by our method that have the maximum score for an original headline, we see that 47 of them were credited as humorous (i.e. having a score $\geq$ 1) out for the 83 original title. On the other hand, 43 of the human generated were considered humorous.

In the following analysis, we aim to evaluate the different criteria considered in our method for modeling humour. In terms of prosody, we look at the number of times a headline was considered to be punny by people with respect to our method's score on the prosody dimension. Overall, 22\% of the generated headlines were considered to have a pun in relation to the original word. Out of these headlines, 88\% of them were evaluated positively on the prosody dimension by our system. This indicates that the method exhibited capability of assessing the sound similarity and punniness to the original word.

For the concreteness, we are considering concrete words defined in \cite{brysbaert2014concreteness} as candidates. As a result, we expected to have headlines produced by the method score high on the fourth question. Contrary, only 56\% of them were deemed concrete. This indicates that a more robust model is required to model the concreteness of terms.

By observing Figure~\ref{fig:dist-plot}, we notice that 46\% of headlines suggested by our method are considered surprising (i.e. scoring at 1, on average). As we are using a word embeddings model, it is difficult to come up with a semantic similarity threshold that separates similar words from non-similar ones, especially for modeling surprisingness. Therefore, we test the scores assigned by the models on three thresholds of similarity (0.3, 0.2 and 0.1) with respect to the headlines viewed as surprising by online people.
Out of the 46\% surprising headlines, 98\%, 84\% and 40\% headlines are considered to be dissimilar by the semantic model by using the three above mentioned thresholds. This indicates that minimizing the semantic similarity increases surprise to a degree, after which lowering the similarity results in a lower surprise.

Lastly, we perform the same analysis regarding the connection between the replacement word and the selected target with respect to question five and six. 75\% of the time, our function scored positively on headlines evaluated as making fun of a target. Out of which, 77\% were correctly seen as negative by the method with respect to Q6.

\section{Discussion}

As the best headlines produced by our system for each original headline can, on the average, reach to a human level in terms of most of the factors measured by our evaluation, an immediate future direction for our research is to develop a better ranking mechanism to reach to the maximum capacity of our system. Perhaps such ranking could be learned by training an LSTM classifier on humor annotated corpora such as the one used in this paper or the one proposed by \cite{west-horvitz-aaai2019-unfun}.

For surprise, we opted for a rather modest approach by assimilating it to an inverse semantic similarity to the original word. However, different metrics have been proposed to model this phenomenon, such as a neural network based composer-audience model \cite{mr.surprise} or probabilistically modelling the likelihood of a certain word occurring in a given context (see \cite{degaetano-ortlieb-piper-2019-scientization}).

The particularly low score on the concreteness highlights the inadequacy of using an annotated lexicon for its assessment. Perhaps, in the future, concreteness could be modelled in a more robust context dependent way. Previous work \cite{naumann2018quantitative} exists showing differences in the distributional representations of concrete and abstract words. As word embedding models are based on the distributional hypothesis, this discovery could be exploited for a context dependent classification by using context-aware word embeddings.

If the method was to be used as a tool for assisting journalists when composing news articles, the fact that employing computational methods for headline generation might result in offensive headlines (see \cite{alnajjar2019no}) has to be taken into account. Our humor model maximizes the negative relation to its target, which might be considered as an insult, if understood in a wrong, non humorous fashion.

Our current approach focuses on English, in the future, we are interested in using our method for other languages as well such as Finnish. This would require a more robust surface realization method to deal with morphology more complex than that of English \cite{hamalainen2018development}. There is already a similar semantic database available for Finnish \cite{hamalainen2018extracting} as the one we used for English, which greatly facilitates a multilingual port of our method.

\section{Conclusions}

We have presented a method for generating humorous headlines that in its current state, falls behind the human level humor. Nevertheless the results reach to a comparable level with an existing neural based method. The method proposed by us has the potential of reaching to a human level humor generation in the limited domain task of altering a word in an existing headline if a better ranking mechanism for its output was introduced. 

The evaluation and analysis we conducted on the results has revealed several features which can be modelled better in the future to improve our method. As we have gathered human judgements for headlines generated by our system for original headlines that are based on an existing humor annotated corpus, we are releasing our evaluation results and the generated titles\footnote{https://zenodo.org/record/4976481} in the same format as the corpus we used so that our data can be easily used in research dealing with the existing dataset.

\section{Acknowledgments}
This work has been partially supported by the European Union's Horizon 2020 programme under grant 825153 (Embeddia).

\bibliographystyle{iccc}
\bibliography{acl2019}

\end{document}